\documentclass[sigconf]{acmart}
\AtBeginDocument{%
  }

\setcopyright{acmlicensed}
\copyrightyear{2018}
\acmYear{2018}
\acmDOI{XXXXXXX.XXXXXXX}
\acmConference[Conference acronym 'XX]{Make sure to enter the correct
  conference title from your rights confirmation email}{June 03--05,
  2018}{Woodstock, NY}
\acmISBN{978-1-4503-XXXX-X/2018/06}

\usepackage{cite}
\usepackage{amsmath,amsfonts}
\usepackage[ruled]{algorithm2e}
\usepackage{textcomp}
\usepackage{multirow}
\usepackage{algorithmic}
\usepackage{colortbl}
\usepackage{hyperref}
\usepackage{cleveref}
\usepackage{caption}
\usepackage{subcaption}
\usepackage{graphicx}

\definecolor{lightgray}{gray}{0.9}
\definecolor{cvprblue}{rgb}{0.21,0.49,0.74}
\makeatletter

\DeclareRobustCommand\onedot{\futurelet\@let@token\@onedot}
\def\@onedot{\ifx\@let@token.\else.\null\fi\xspace}
\def\eg{\emph{e.g}\onedot}

\makeatother

\begin{document}

\title{Pseudo Multi-Source Domain Generalization: Bridging the Gap Between Single and Multi-Source Domain Generalization}

\author{Shohei Enomoto}
\email{shohei.enomoto@ntt.com}
\orcid{0009-0000-7450-3311}
\affiliation{%
  \institution{NTT}
  \city{Musashino}
  \state{Tokyo}
  \country{Japan}
}

\renewcommand{\shortauthors}{Enomoto}

\begin{abstract}
Deep learning models often struggle to maintain performance when deployed on data distributions different from their training data, particularly in real-world applications where environmental conditions frequently change. 
While Multi-source Domain Generalization (MDG) has shown promise in addressing this challenge by leveraging multiple source domains during training, its practical application is limited by the significant costs and difficulties associated with creating multi-domain datasets. 
To address this limitation, we propose Pseudo Multi-source Domain Generalization (PMDG), a novel framework that enables the application of sophisticated MDG algorithms in more practical Single-source Domain Generalization (SDG) settings. 
PMDG generates multiple pseudo-domains from a single source domain through style transfer and data augmentation techniques, creating a synthetic multi-domain dataset that can be used with existing MDG algorithms. 
Through extensive experiments with PseudoDomainBed, our modified version of the DomainBed benchmark, we analyze the effectiveness of PMDG across multiple datasets and architectures. 
Our analysis reveals several key findings, including a positive correlation between MDG and PMDG performance and the potential of pseudo-domains to match or exceed actual multi-domain performance with sufficient data. 
These comprehensive empirical results provide valuable insights for future research in domain generalization. 
Our code is available at \url{https://github.com/s-enmt/PseudoDomainBed}.
\end{abstract}




\received{20 February 2007}
\received[revised]{12 March 2009}
\received[accepted]{5 June 2009}

\maketitle

\section{Introduction}
Deep learning models have achieved remarkable success across various domains.
However, their performance often deteriorates significantly when tested on data distributions different from their training data.
This challenge is particularly prevalent in outdoor applications such as autonomous driving and smart cities, where environmental factors like weather conditions and lighting variations can substantially alter the input distribution.
Therefore, developing robust deep learning models that maintain their performance on unseen distributions is crucial for real-world applications.

Domain generalization (DG) has emerged as a promising approach to address this challenge. 
DG techniques can be broadly categorized into two settings: Single-source Domain Generalization (SDG), which uses data from a single source domain, and Multi-source Domain Generalization (MDG), which leverages data from multiple source domains. 
While most existing research focuses on the MDG setting using multi-domain datasets (\eg, PACS dataset~\citep{pacs} comprising Photo, Art, Cartoon, and Sketch domains), creating such datasets is often impractical due to the high costs associated with data collection and annotation. 
This limitation significantly hinders the practical application of MDG algorithms.

To bridge this gap, we propose Pseudo Multi-source Domain Generalization (PMDG), a novel framework that enables the application of sophisticated MDG algorithms in more practical SDG settings.
Our approach generates multiple pseudo-domains from a single source domain, treating them as distinct domains to create a synthetic multi-domain dataset.
We investigate two approaches for effective pseudo-domain generation, style transformation and data augmentation.
For style transformation, inspired by the PACS dataset, we employ AdaIN Style Transfer~\citep{adain_style_transfer,sty-in}, CartoonGAN~\citep{cartoongan}, and Edge Detection~\citep{dexined,dexined_journal} to generate Art-style, Cartoon-style, and Sketch-style images.
For data augmentation, we utilize both conventional augmentation techniques and robust augmentation methods.

\begin{figure}[tb!]
\centering
\includegraphics[width=1.0\linewidth]{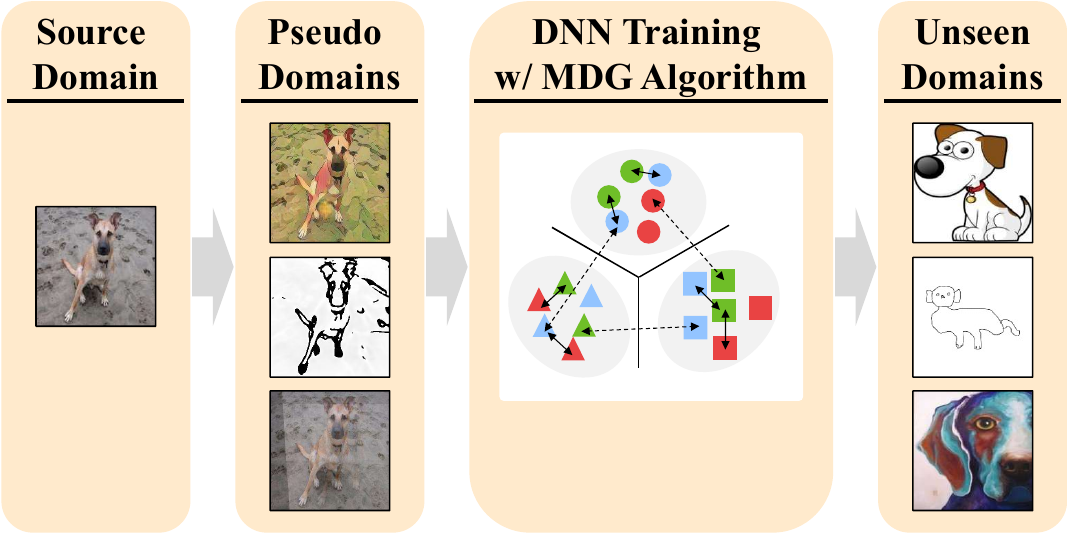}
\caption{
Overview of PMDG framework.
PMDG applies multiple transformations to training samples to generate pseudo-domains.
The DNN is then trained using an MDG algorithm on these pseudo-domains.
The trained DNN aims to be robust against unknown domains.
}
\label{fig:PMDG_overview}
\Description{
Overview of PMDG framework.
PMDG applies multiple transformations to training samples to generate pseudo-domains.
The DNN is then trained using an MDG algorithm on these pseudo-domains.
The trained DNN aims to be robust against unknown domains.
}
\end{figure}

To evaluate PMDG, we create PseudoDomainBed, a modified version of the popular MDG benchmark DomainBed~\citep{domainbed}, adapted for the SDG setting. 
Through extensive experiments on multiple datasets and architectures (ResNet50 and ViT), we demonstrate that PMDG achieves superior performance compared to existing SDG algorithms. 
Our analysis reveals several key findings, including a positive correlation between MDG and PMDG performance, the potential of pseudo-domains to match or exceed actual multi-domain performance, and the importance of architecture-specific pseudo-domain generation strategies.

Our main contributions are as follows:
\begin{itemize}
\item We propose PMDG, a novel framework that bridges the gap between MDG and SDG, enabling the application of sophisticated MDG algorithms in practical SDG settings.
\item We introduce PseudoDomainBed, a modified version of DomainBed for fair comparison and evaluation, with publicly available code. PseudoDomainBed facilitates easy utilization of MDG algorithms implemented in DomainBed and enables their evaluation in SDG settings through pseudo-domains.
\item Through extensive experiments, we demonstrate that PMDG outperforms existing SDG methods and provide valuable insights to motivate future domain generalization research.
\end{itemize}

Looking ahead, our evaluation opens up several promising directions for future research, particularly in understanding when and why pseudo-domains can effectively substitute for actual domains.

\section{Related Work}
This section discusses domain generalization research, which has traditionally been divided into two distinct approaches: Single-source Domain Generalization (SDG) and Multi-source Domain Generalization (MDG).
Although these approaches have developed independently, our work aims to bridge the gap between them, enabling cross-pollination of ideas and methodologies.

\subsection{Multi-source Domain Generalization (MDG)}
The majority of domain generalization research operates under the MDG paradigm, which assumes access to training data from multiple domains with a shared label space.
MDG approaches aim to learn domain-invariant information to improve generalization to unseen domains. 
Researchers have proposed various approaches including learning domain-invariant features~\citep{dann,mmd,coral,irm,csa}, regularization techniques~\citep{dro,rsc,vrex,selfreg,arm,fish,ridg,sd,swad}, data augmentation strategies~\citep{inter_mixup,sagnet,rsc,jigen}, self-supervised learning~\citep{jigen,pden,selfreg}, causal perspective~\citep{irm,vrex}, meta-learning~\citep{mldg}, and architectural innovations~\citep{gmoe}.

While these approaches have shown promising results, their practical applicability has been limited by the requirement of multi-domain training data. 
Our proposed PMDG framework addresses this limitation by enabling these sophisticated MDG algorithms to operate effectively in more practical SDG settings.

\subsection{Single-source Domain Generalization (SDG)}
SDG research focuses on achieving domain generalization using training data from a single domain, which better reflects real-world scenarios. 
This approach is particularly relevant given that many widely-used computer vision datasets (\eg, ImageNet~\citep{imagenet}) consist of data from a single domain.
Current SDG approaches can be broadly categorized into three groups: learning algorithms, domain expansion methods, and data augmentation techniques.
Learning algorithms~\citep{rsc,sagnet,sd} aim to prevent overfitting to source domain-specific information by introducing additional training objectives.
Domain expansion methods systematically generate novel domains through various approaches: domain generator~\citep{wang2021learning,pden}, uncertainty-guided generation~\citep{qiao2021uncertainty}, optimal transport~\citep{zhou2020learning}, and adversarial data augmentation~\citep{qiao2020learning,volpi2018generalizing,zhao2020maximum}.
Data augmentation methods~\citep{augmix,ipmix,pixmix,augmax,vaish2024fourier,randconv,pro_randconv,na2021fixbi,9878911,mixstyle} focus on increasing the diversity of training data through various transformations, often accompanied by specialized training procedures such as custom loss functions.

While domain expansion methods offer sophisticated domain generation techniques, they often require complex adversarial attacks or specialized architectures that can be unstable and computationally intensive. 
In contrast, data augmentation methods provide a more straightforward and computationally efficient approach to creating pseudo-domains, making them better suited for our goal of benchmarking MDG algorithms. 
% Furthermore, data augmentation techniques have proven effective at capturing real-world variations in computer vision tasks, aligning well with our objective of simulating domain shifts.

We treat augmented data as samples from different domains and utilize data augmentation techniques for pseudo-domain generation.
Since identifying effective transformations and their combinations for pseudo-domain generation is non-trivial, we conduct extensive empirical studies to address these questions.
Our results demonstrate that pseudo-domains can serve as a practical testbed for the rich collection of MDG algorithms, suggesting that future research efforts should focus on developing effective pseudo-domain generation strategies rather than specialized learning algorithms.

\section{SDG Problem Setting}
Our research follows the SDG problem setting.
In SDG, we aim to learn a model that can generalize to unknown target domains using only a single source domain. Let us formally define this problem setting.

Let $D^{S} = \{(x_i, y_i)\}_{i=1}^n$ be a source domain dataset, where $x_i \in \mathcal{X}$ represents input data, $y_i \in \mathcal{Y}$ represents corresponding labels, and $(x_i, y_i)$ follows the source domain distribution $P^{S}(X,Y)$.

We consider a set of unknown target domains $\mathcal{T} = \{T_1, T_2, ..., T_k\}$, where each target domain $T_j$ has a different distribution from the source domain:
\begin{equation}
   P^{T_j}(X,Y) \neq P^{S}(X,Y)
\end{equation}

The objective of SDG is to learn a function $f_{\theta}: \mathcal{X} \to \mathcal{Y}$ that minimizes the expected risk across all potential target domains:
\begin{equation}
   f_{\theta}^* = \arg\min_{f_{\theta}} \mathbb{E}_{T_j \in \mathcal{T}} [\mathbb{E}_{(x,y) \sim P^{T_j}}[L(f_{\theta}(x), y)]]
\end{equation}
where $L: \mathcal{Y} \times \mathcal{Y} \to \mathbb{R}$ is a loss function.

% The key challenge in SDG lies in three fundamental constraints. First, only the source domain $S$ is available during training. Second, no access to any target domain data is permitted during training. Third, there exists a significant distribution shift between domains that must be addressed.

% This setting differs from traditional domain adaptation and multi-domain generalization in that it requires learning domain-invariant features from a single source domain, making it more challenging yet more practical for real-world applications where collecting multi-domain datasets is difficult.

\section{Proposed Framework}
We propose Pseudo Multi-source Domain Generalization (PMDG), a novel framework that enables the application of MDG algorithms to single-source datasets by generating pseudo multi-domain datasets through various transformations. 
\Cref{fig:PMDG_overview} provides an overview of PMDG, while \Cref{alg:PMDG} details its implementation.

\begin{algorithm}
\centering
\caption{Training a DNN with PMDG}
\label{alg:PMDG}
\begin{algorithmic}[1]
    \STATE \textbf{Input:} Training dataset $D^{S} = \{(x_i, y_i)\}_{i=1}^n$ , \\
    Number of epochs $ E $, Batch size $ B $, Transformations $\mathcal{O}=\{ O_1, O_2, \dots, O_N \}$
    \STATE \textbf{Initialize} model parameters $ \theta $
    \FOR{epoch $ e = 1 $ to $ E $}
        \STATE Shuffle the training dataset $D^{S}$
        \FOR{each mini-batch $ \{(x_b, y_b)\}_{b=1}^B $ in $ D^{S}$}
            \STATE Generate pseudo-domains: $\{\tilde{x}_b^k = O_k(x_b)\}_{k=1}^K$ for each image in mini-batch
            \STATE Obtain predictions: $\{\hat{y}_b^k = f_\theta(\tilde{x}_b^k)\}_{k=1}^K$ for each pseudo-domains
            \STATE Compute MDG loss: $ L^{\mathrm{MDG}}$ using algorithm-specific objectives
            \STATE Update model parameters $ \theta \leftarrow \theta - \eta \nabla_\theta L^{\mathrm{MDG}} $
        \ENDFOR
    \ENDFOR
    \STATE \textbf{Output:} Trained model parameters $ \theta $
\end{algorithmic}
\end{algorithm}

\subsection{Pseudo-Domain Generation}
Identifying effective methods for generating pseudo-domains remains a crucial open challenge.
In this research, we introduce and evaluate two approaches, both independently and in combination, style transformation and data augmentation.

\subsubsection{Style Transformation}
Inspired by the PACS dataset, we propose three transformations to recreate its constituent domains.
The first transformation is AdaIN style transfer~\citep{adain_style_transfer,sty-in}, a technique that preserves image content (shape) while modifying style, used for creating art-style images.
The second is CartoonGAN~\citep{cartoongan}, a GAN-based approach for transforming images into cartoon-style representations. 
The third is Edge Detection~\citep{dexined,dexined_journal}, a method for extracting image contours used to generate sketch-style images. 
The visual results of these transformations are shown in \Cref{fig:visualize}.

\subsubsection{Data Augmentation}
\begin{figure*}[tb!]
\centering
\resizebox{\textwidth}{!}{
\begin{tabular}{cccccc}
Original & MixUp & CutMix & RandAugment & TrivialAugment & AugMix  \\
\includegraphics[width=0.15\linewidth]{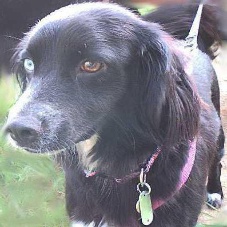} &
 \includegraphics[width=0.15\linewidth]{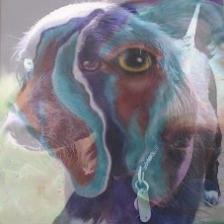}  & \includegraphics[width=0.15\linewidth]{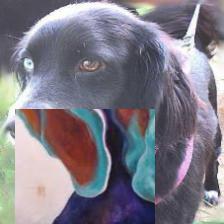}  & \includegraphics[width=0.15\linewidth]{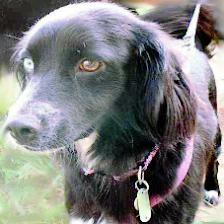}  &  \includegraphics[width=0.15\linewidth]{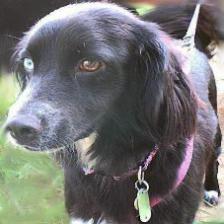}&
 \includegraphics[width=0.15\linewidth]{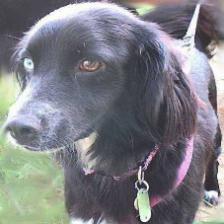}    \\
 IPMix & RandConv & Style Transfer & CartoonGAN & Edge Detection  \\
\includegraphics[width=0.15\linewidth]{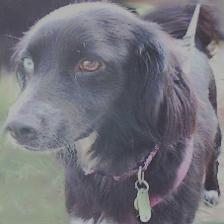} &
\includegraphics[width=0.15\linewidth]{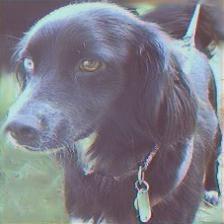} &
\includegraphics[width=0.15\linewidth]{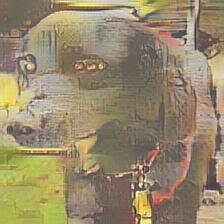} &
\includegraphics[width=0.15\linewidth]{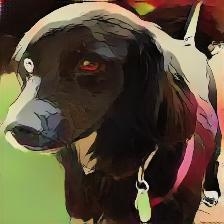} &
\includegraphics[width=0.15\linewidth]{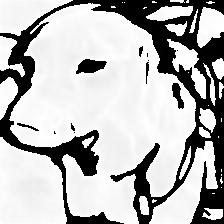}   
\end{tabular}
}
\caption{
Visualization of the transformed sample.
We performed different types of transformations on the dog images of the PACS dataset.
}
\label{fig:visualize}
\Description{
Visualization of the transformed sample.
We performed different types of transformations on the dog images of the PACS dataset.
}
\end{figure*}

We employ various data augmentation techniques of different types.
These include mixing-based methods such as MixUp~\citep{mixup} and CutMix~\citep{cutmix}, advanced augmentation strategies including RandAugment~\citep{randaugment} and TrivialAugment~\citep{trivialaugment}, and robustness-focused augmentations comprising AugMix~\citep{augmix}, IPMix~\citep{ipmix}, and RandConv~\citep{randconv}. 
While some augmentation techniques have associated loss functions, we omit them for simplicity in this study. 
Visual examples are presented in \Cref{fig:visualize}.

\subsubsection{Transformation Selection}
Given the limited understanding of optimal transformation count and inter-transformation interactions, we take an empirical approach. 
We select $K$ transformations (with replacement) from a predefined set of transformation operations to construct the transformation set $\mathcal{O}=\{O_1, O_2, \dots, O_K\}$, where each $O$ represents an individual transformation operation.

\subsubsection{Transformation Application}
We generate pseudo multi-domain data using the transformation set $\mathcal{O}$. 
Specifically, for an input mini-batch $B = {(x_i, y_i)}_{i=1}^b$, we apply each transformation in the set to obtain $K$ pseudo multi-domain mini-batches $\{B_1, \dots, B_K\}$:

\begin{equation}
B_k = O_k(B), \quad k = 1,\dots,K
\end{equation}

\subsection{Training with MDG Algorithm}
We train the model $f_{\theta}$ using an MDG algorithm on the $K$ pseudo multi-domain mini-batches. 
While various MDG algorithms have been proposed---including feature alignment, regularization, and meta-learning approaches---our framework is algorithm-agnostic and can accommodate any MDG algorithm.

\section{Experimental Setup}
We evaluate our approach using standard domain generalization datasets. 
Since model selection significantly impacts performance evaluation in domain generalization, we ensure fair comparison by modifying DomainBed~\citep{domainbed}, the standard MDG benchmark, to accommodate the SDG setting. 
We call our modified benchmark PseudoDomainBed, which implements our pseudo-domain generation approach.

% \subsection{Implementation Details of PseudoDomainBed}
% Here we detail the key differences between DomainBed and our PseudoDomainBed implementation.

% \textbf{Training Configuration.}
% Following the original implementation, we maintain consistent training configurations with DomainBed, including learning rate, batch size, and other hyperparameters.

% \textbf{Architecture.}
% We use ResNet50~\citep{resnet} as our backbone network. 
% Following common practice in single-domain settings, we maintain the batch normalization layers in the network.

% \textbf{Pseudo-Domain Generation.}
% Our pseudo-domain transformations are implemented as data augmentations applied at the mini-batch level. 
% We use default hyperparameters for all transformation operations as reported in their respective papers.

% \textbf{Model Selection.}
% We employ training-domain validation sets for model selection, as this approach has shown the most stable performance in MDG settings.

% \textbf{Data Processing Pipeline.}
% The pseudo-domain transformations are applied after the default data augmentation pipeline of DomainBed.
% Notably, while the original DomainBed applies data augmentation to validation sets, we omit this in PseudoDomainBed to avoid distorting the evaluation of generalization to unknown distributions.
\subsection{Implementation Details of PseudoDomainBed}
\subsubsection{Key differences between DomainBed and PseudoDomainBed}
% Here we detail the key differences between DomainBed and our PseudoDomainBed implementation.
Following the original implementation, we maintain consistent training configurations with DomainBed, including learning rate, batch size, and other hyperparameters. 
We use ResNet50~\citep{resnet} as our backbone network, maintaining the batch normalization layers as per common practice in single-domain settings. 
For model selection, we employ training-domain validation sets, which has shown the most stable performance in MDG settings.
Our pseudo-domain transformations are implemented as data augmentations applied at the mini-batch level, using default hyperparameters from their respective papers. 
These transformations are applied after the default data augmentation pipeline of DomainBed. 
Notably, while the original DomainBed applies data augmentation to validation sets, we omit this in PseudoDomainBed to avoid distorting the evaluation of generalization to unknown distributions.

\subsubsection{Pseudo-domain Generation}
Implementing various transformation techniques in a unified framework presents challenges due to their different operating levels. 
To address this, we carefully designed the implementation architecture of PseudoDomainBed to handle different types of transformations consistently.
Image transformations in our framework can be categorized into two levels based on their processing stage.
The first category is dataset-level transformations, which operate on raw images before converting them to tensors. 
Transformations such as RandAugment fall into this category, where augmentations are applied directly to image data.
The second category is mini-batch-level transformations, which operate on normalized tensors during the training process. 
For example, MixUp belongs to this category as it combines multiple normalized image tensors.
To handle these different transformation types uniformly, we implemented a standardized interface for pseudo-domain generation algorithms. 
Each algorithm is required to implement both dataset-level and mini-batch-level transformation methods, even if only one is actually used. 
This design choice provides a consistent API for users to employ any transformation technique without considering its implementation level. 
It also enables flexible integration of new transformation methods by implementing the standard interface. 

\subsection{Datasets}
Our primary evaluation uses four standard domain generalization datasets: PACS~\citep{pacs}, VLCS~\citep{vlcs}, OfficeHome~\citep{officehome}, and TerraIncognita~\citep{terraincognita}. 
% Additionally, we provide results on Wilds datasets~\citep{wilds} to demonstrate real-world applications of domain generalization.

\subsection{SDG Baselines}
We compare our method against standard SDG baselines: ERM~\citep{erm}, Mixup~\citep{mixup}, SagNet~\citep{sagnet}, RSC~\citep{rsc}, AugMix~\citep{augmix}, CutMix~\citep{cutmix}, IPMix~\citep{ipmix}, RandAugment~\citep{randaugment}, RandConv~\citep{randconv} and TrivialAugment~\citep{trivialaugment}. 
These baselines represent the current state-of-the-art approaches in single-domain generalization.

\subsection{MDG Algorithms}
We utilize the MDG algorithms implemented in DomainBed with their default hyperparameters. 
Specifically, we evaluate our approach with algorithms ARM~\citep{arm}, CDANN~\citep{cdann}, CORAL~\citep{coral}, DANN~\citep{dann}, EQRM~\citep{eqrm}, GroupDRO~\citep{dro}, IRM~\citep{irm}, MLDG~\citep{mldg}, MMD~\citep{mmd}, MTL~\citep{mtl}, RIDG~\citep{ridg}, SD~\citep{sd}, SelfReg~\citep{selfreg} and VREx~\citep{vrex}. 
% Note that some MDG algorithms requiring modifications beyond algorithm.py are not directly applicable in our setting.

\section{Results}
We report the mean and standard error over three trials for each experiment conducted on PseudoDomainBed. 
Our evaluation demonstrates that PMDG outperforms SDG baselines.

% \subsection{Main Results}
% We evaluated the effectiveness of using style transfers and data augmentations as pseudo-domains, both individually and in combination. The source domain without any transformations was always included in the training process. All experiments used ResNet50 with batch normalization layers, and we employed ERM and SD as the MDG algorithms.
% Table 1 presents our comprehensive results. Data augmentation methods show varying levels of effectiveness, with IPMix demonstrating the strongest performance among all augmentation techniques tested. Style transfer methods exhibit [specific performance characteristics].
% In our combination studies, we found that combining multiple data augmentation methods generally yields limited improvements. However, multiple applications of IPMix show substantial benefits. The most effective combination proved to be style transfer with IPMix, suggesting a complementary relationship between these approaches.
\subsection{Evaluation of Pseudo-domain Generation}
\begin{figure*}[tb!]
\centering
\includegraphics[width=1.0\linewidth]{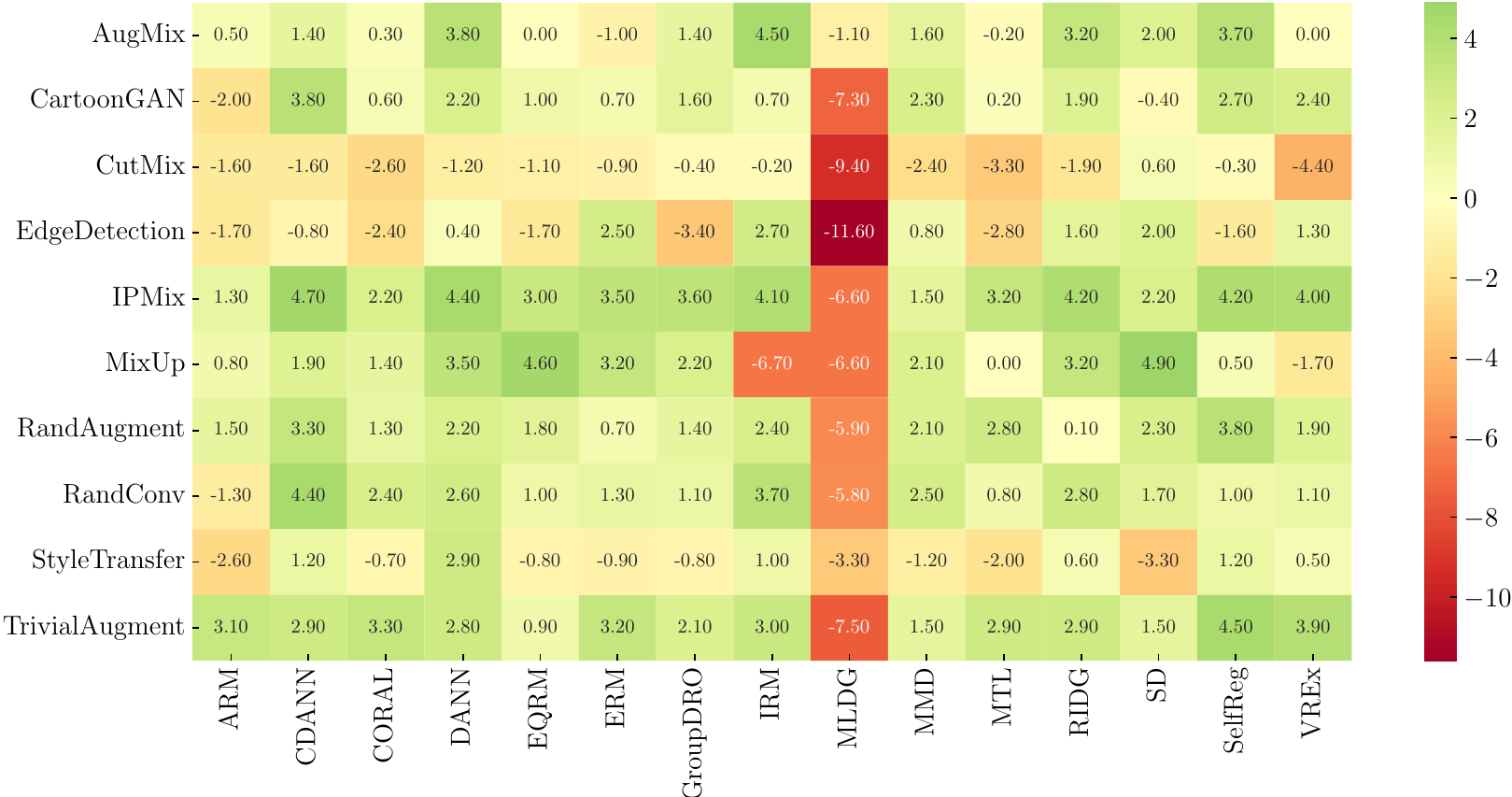}
\caption{
Accuracy gains over the ERM baseline without pseudo-domain across different transformation techniques (y-axis) and MDG algorithms (x-axis) on the VLCS dataset. 
Green and red colors indicate performance improvements and degradation, respectively. 
Values represent accuracy differences from ERM without pseudo-domains.
}
\label{fig:gain_vlcs}
\Description{}
\end{figure*}

We first evaluated various transformation techniques for pseudo-domain generation in combination with different MDG algorithms on the VLCS dataset.   
In the experiment, we consider a two-domain setting consisting of the source domain and one pseudo-domain. 
To assess the effectiveness of each combination, we measure the accuracy gains from the ERM baseline.
\Cref{fig:gain_vlcs} shows a heatmap visualization of these results.
IPMix, RandConv, and TrivialAugment show positive accuracy gains with most MDG algorithms, suggesting their effectiveness as pseudo-domain generation techniques. 
In contrast, CutMix leads to performance degradation in most cases. 
Notably, MLDG shows performance deterioration across all transformation techniques, suggesting its incompatibility with our pseudo-domain approach.
The negative results with MLDG suggest that not all MDG algorithms are suitable for pseudo-domain settings, possibly due to their underlying assumptions about domain characteristics.

\subsection{Analysis of Pseudo-domain Combinations}
We evaluated two different pseudo-domain combinations with various MDG algorithms. 
The first combination consists of three domains Org+IM+IM, while the second combination includes six domains Org+ST+ED+CG+IM+IM. 
We compare these combinations using three learning algorithms: ERM, RIDG, and SD.
\Cref{tab:main} presents the results. 
The combination of SD with Org+IM+IM achieves the highest accuracy of $55.9\%$, surpassing the best SDG baseline, IPMix ($55.2\%$). 
Interestingly, while the addition of style-based transformations (ST, ED, and CG) leads to significant improvements on the PACS dataset, its effectiveness is limited on other datasets.
Evaluating methods solely on PACS might lead to the development of techniques that excel only on style-based domain shifts while failing to generalize to other types of distribution shifts. 
This observation emphasizes the importance of diverse evaluation protocols that include multiple datasets with different types of domain shifts.
While specialized techniques like style-based transformations can be highly effective for specific scenarios, they should be complemented by more general-purpose approaches to ensure broader applicability.

% \subsection{Performance on Vision Transformers}
Additionally, to investigate the effectiveness of our PMDG framework across different architectures, we conducted similar experiments using Vision Transformer (ViT)~\citep{vit,deit} as the backbone network. 
\Cref{tab:main} presents the results.
The combination of SD with pseudo-domains Org+ST+ED+CG+IM+IM achieves the highest accuracy of $62.0\%$, surpassing the best SDG baseline ($60.5\%$). 
Style-based transformations that enhance shape features appear to be more effective with ViT~\citep{shape_texture,vit_shape_texture}, suggesting that the choice of pseudo-domain generation techniques should consider the underlying architectural characteristics of the backbone network.

\begin{table*}[tb!]
\centering
\caption{
Accuracy comparison on four datasets using different combinations of algorithms and pseudo-domains. 
``Avg'' represents the mean accuracy across all datasets. 
The upper and lower tables show results with ResNet50 and ViT backbones, respectively.
In each table, the upper part shows SDG baseline results and the lower part presents PMDG results. 
\textbf{Bold} and \underline{underlined} numbers denote the first and second highest accuracy, respectively. 
Org denotes the original domain without transformation, IM represents pseudo-domains generated by IPMix, ST by StyleTransfer, ED by EdgeDetection, and CG by CartoonGAN.
$^\dagger$ indicates exclusion of domain-specific transformations during training: ST is excluded when testing on Art domain, CG for Cartoon domain, and ED for Sketch domain.
}
\label{tab:main}
% \resizebox{\textwidth}{!}{
\begin{tabular}{llccccc}
\toprule
\textbf{Algorithm}        & \textbf{Pseudo-domain}        & \textbf{VLCS}             & \textbf{PACS}             & \textbf{OfficeHome}       & \textbf{TerraIncognita}   & \textbf{Avg}              \\
\midrule
ERM    &  --                   & 60.6 $\pm$ 1.3            & 56.3 $\pm$ 0.5            & 53.4 $\pm$ 0.1            & 32.2 $\pm$ 0.7            & 50.6                      \\
Mixup      &  --               & 64.3 $\pm$ 1.2            & 55.9 $\pm$ 0.8            & 54.5 $\pm$ 0.3            & 32.7 $\pm$ 0.5            & 51.8                      \\
SagNet    &  --                & 62.0 $\pm$ 0.3            & 52.2 $\pm$ 0.4            & 51.5 $\pm$ 0.3            & 32.6 $\pm$ 0.7            & 49.6                      \\
RSC         &  --              & \underline{65.0} $\pm$ 0.9 & 56.1 $\pm$ 1.3            & 52.5 $\pm$ 0.3            & 33.3 $\pm$ 0.2            & 51.7                      \\
AugMix        &  --            & 61.8 $\pm$ 0.5            & 57.8 $\pm$ 0.4            & 54.6 $\pm$ 0.3            & 32.1 $\pm$ 0.3            & 51.6                      \\
CutMix         &  --           & 61.8 $\pm$ 0.3            & 54.7 $\pm$ 1.0            & 53.9 $\pm$ 0.1            & 33.0 $\pm$ 1.1            & 50.8                      \\
IPMix       &  --              & 64.6 $\pm$ 1.0            & 65.9 $\pm$ 0.3            & \underline{55.6} $\pm$ 0.2            & 34.9 $\pm$ 0.7            & 55.2                      \\
RandAugment     &  --          & 58.6 $\pm$ 0.8            & 58.9 $\pm$ 1.0            & 53.9 $\pm$ 0.2            & 33.2 $\pm$ 0.5            & 51.1                      \\
RandConv         &  --         & 62.1 $\pm$ 0.1            & 62.8 $\pm$ 0.7            & 53.2 $\pm$ 0.3            & 34.7 $\pm$ 0.3            & 53.2                      \\
TrivialAugment     &  --       & 61.1 $\pm$ 1.1            & 59.9 $\pm$ 1.5            & 54.2 $\pm$ 0.2            & 36.2 $\pm$ 0.2            & 52.8                      \\
\midrule
\rowcolor{lightgray}ERM & Org+IM+IM                        & 64.6 $\pm$ 1.5            & 63.4 $\pm$ 1.3            & 55.1 $\pm$ 0.4            & \underline{36.6 $\pm$ 0.9}            & 54.9                      \\
\rowcolor{lightgray}ERM & Org+ST+ED+CG+IM+IM$^\dagger$     & 64.9 $\pm$ 1.1            & 69.9 $\pm$ 0.5            & 55.4 $\pm$ 0.1                     & 31.1 $\pm$ 0.6                        & \underline{55.3}                      \\
\rowcolor{lightgray}RIDG & Org+IM+IM                       & 63.4 $\pm$ 1.6            & 64.8 $\pm$ 0.3            & 55.4 $\pm$ 0.5            & 37.2 $\pm$ 0.2                        & 55.2                      \\
\rowcolor{lightgray}RIDG & Org+ST+ED+CG+IM+IM$^\dagger$    & 61.7 $\pm$ 0.0            & \textbf{71.8} $\pm$ 0.4             & 55.2 $\pm$ 0.2                      & 31.7 $\pm$ 0.4                        & 55.1                      \\
\rowcolor{lightgray}SD & Org+IM+IM                         & \textbf{65.6} $\pm$ 1.2   & 64.1 $\pm$ 1.0            & \textbf{56.7} $\pm$ 0.2   & \textbf{37.1 $\pm$ 0.7}               & \textbf{55.9}    \\
\rowcolor{lightgray}SD  & Org+ST+ED+CG+IM+IM$^\dagger$    & 61.4 $\pm$ 0.3             & \underline{69.7} $\pm$ 0.4         & \underline{55.6} $\pm$ 0.2          & 30.5 $\pm$ 0.9                        & 54.3                     \\
\bottomrule
\end{tabular}
% }
% \end{table*}

% \begin{table*}[tb!]
% \centering
% \caption{
% Accuracy comparison on four datasets using different combinations of algorithms and pseudo-domains with ViT. 
% ``Avg'' represents the mean accuracy across all datasets. 
% The upper part shows SDG baseline results, while the lower part presents PMDG results. 
% \textbf{Bold} and \underline{underlined} numbers denote the first and second highest accuracy, respectively. 
% $^\dagger$ indicates exclusion of domain-specific transformations during training: ST is excluded when testing on Art domain, CG for Cartoon domain, and ED for Sketch domain.
% }
% \label{tab:deit}
% \resizebox{\textwidth}{!}{
\begin{tabular}{llccccc}
\\
\toprule
\textbf{Algorithm}   & \textbf{Pseudo-domain}            & \textbf{VLCS}                               & \textbf{PACS}                 & \textbf{OfficeHome}                         & \textbf{TerraIncognita}                     & \textbf{Avg}                      \\ \midrule
ERM   & --                           & 64.5 $\pm$ 0.8                              & 73.7 $\pm$ 0.5                & 66.7 $\pm$ 0.4                              & 32.1 $\pm$ 0.8                              & 59.3                              \\
IPMix  & --                          & 65.8 $\pm$ 1.0                              & 76.3 $\pm$ 1.0                & 66.2 $\pm$ 0.3                              & 33.9 $\pm$ 0.1                              & 60.5                              \\ \midrule
\rowcolor{lightgray}ERM & Org+IM+IM                     & \textbf{67.8} $\pm$ 0.1    & 74.5 $\pm$ 1.2                & 67.3 $\pm$ 0.2                              & 33.7 $\pm$ 0.2                              & 60.8                              \\
\rowcolor{lightgray}ERM & Org+ST+ED+CG+IM+IM$^\dagger$  & 66.4 $\pm$ 0.6                              & 79.6 $\pm$ 0.3                          & 66.8 $\pm$ 0.1                                        & 31.3 $\pm$ 0.4                              & 61.0                              \\
\rowcolor{lightgray}RIDG & Org+IM+IM                    & \underline{67.6} $\pm$ 0.3 & 76.1 $\pm$ 0.8                & 68.1 $\pm$ 0.4                              & \textbf{35.7} $\pm$ 0.7    & \underline{61.9} \\
\rowcolor{lightgray}RIDG & Org+ST+ED+CG+IM+IM$^\dagger$ & 64.4 $\pm$ 0.3                              & \underline{80.8} $\pm$ 0.3 & 67.9 $\pm$ 0.2                                        & 32.0 $\pm$ 0.5                              & 61.3                              \\
\rowcolor{lightgray}SD & Org+IM+IM                      & 67.5 $\pm$ 1.8                              & 76.7 $\pm$ 0.6                & \underline{68.4} $\pm$ 0.1 & \underline{34.3} $\pm$ 0.3 & 61.7                              \\
\rowcolor{lightgray}SD & Org+ST+ED+CG+IM+IM$^\dagger$   & 66.4 $\pm$ 0.7                              & \textbf{81.3} $\pm$ 0.3                 & \textbf{68.6} $\pm$ 0.1                               & 31.7 $\pm$ 0.8                              & \textbf{62.0}                     \\ \bottomrule
\end{tabular}
% }
\end{table*}

\subsection{Correlation with MDG Performance}
\begin{figure}[tb!]
\centering
\includegraphics[width=1.0\linewidth]{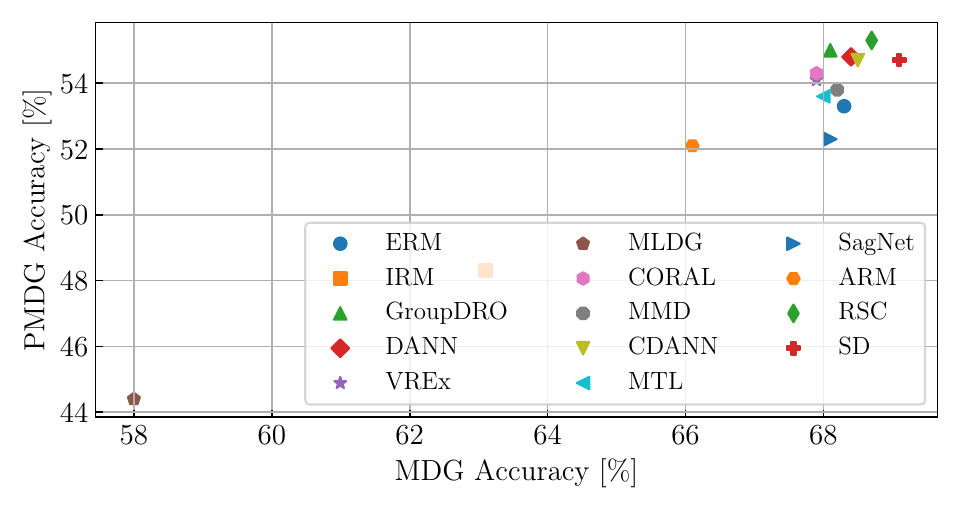}
\caption{
Accuracy comparison of MDG algorithms across MDG and PMDG settings. 
Each point represents a different MDG algorithm.
Accuracy represents averages across PACS, VLCS, OfficeHome, and TerraIncognita datasets.
}
\label{fig:Acc_MDGvsPMDG}
\Description{}
\end{figure}

To understand which MDG algorithms are most suitable for PMDG, we investigated whether the optimal choice of MDG algorithms differs between MDG and PMDG settings, as the training domains in these settings are fundamentally different (real domains vs. pseudo-domains). 
For simplicity, we used Org+IM as our pseudo-domain configuration.
\Cref{fig:Acc_MDGvsPMDG} visualizes the relationship between algorithm performance in MDG and PMDG settings.
The results show a positive correlation between MDG and PMDG accuracy. 
This correlation has two important implications. 
First, it suggests that MDG algorithms that perform well in traditional multi-domain settings are also well-suited for our pseudo-domain approach.
Second, it suggests that the PMDG framework could enhance the practical utility of MDG algorithms by enabling their application to single-domain problems.

\subsection{Quality Assessment: PMDG Dataset vs. Actual Multi-source Domain Dataset}
\begin{figure*}[tb!]
\centering
\begin{subfigure}[tb!]{0.45\textwidth}
    \includegraphics[width=\textwidth]{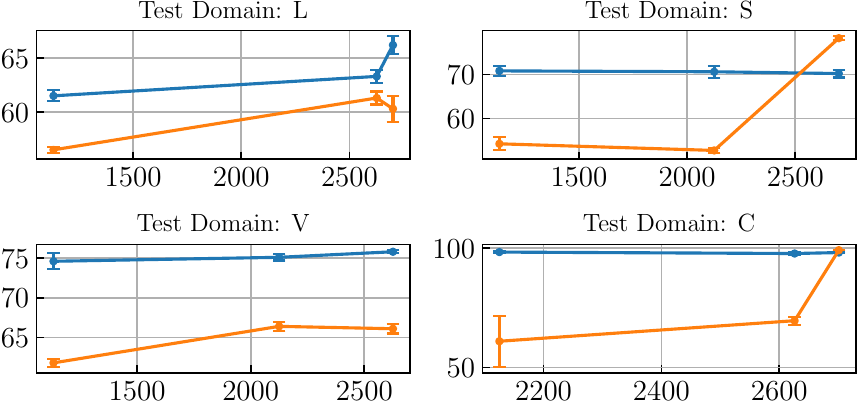}
    \caption{
    VLCS dataset. 
    % Each subplot corresponds to a different test domain: LabelMe (top-left), VOC2007 (bottom-left), SUN09 (top-right), and Caltech101 (bottom-right).
    }
\end{subfigure}
\begin{subfigure}[tb!]{0.45\textwidth}
    \includegraphics[width=\textwidth]{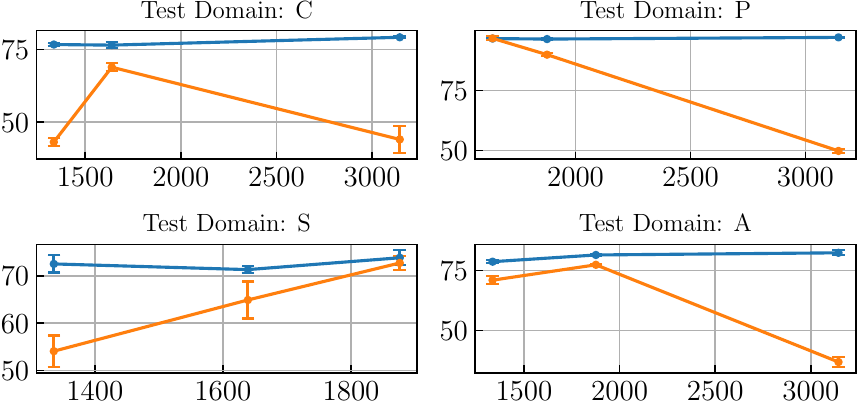}
    \caption{
    PACS dataset. 
    % Each subplot corresponds to a different test domain: Cartoon (top-left), Sketch (bottom-left), Photo (top-right), and Art Painting (bottom-right). 
    }
\end{subfigure}

\begin{subfigure}[tb!]{0.45\textwidth}
    \includegraphics[width=\textwidth]{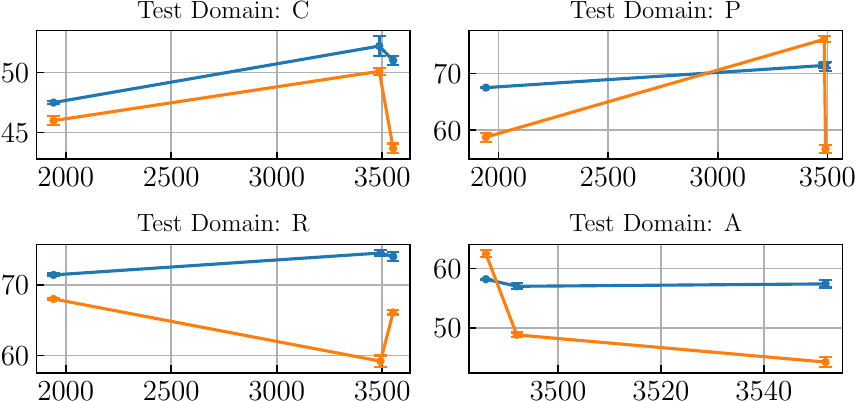}
    \caption{
    OfficeHome dataset. 
    % Each subplot corresponds to a different test domain: Clipart (top-left), Real-World (bottom-left), Product (top-right), and Art (bottom-right). 
    }
\end{subfigure}
\begin{subfigure}[tb!]{0.45\textwidth}
    \includegraphics[width=\textwidth]{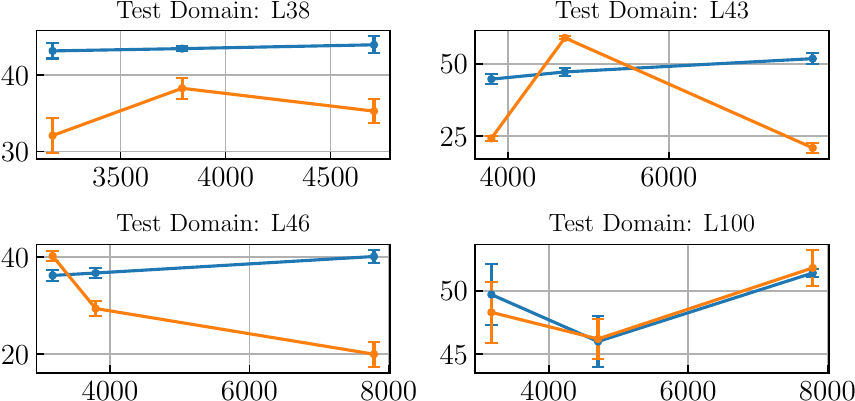}
    \caption{
    TerraIncognita dataset. 
    % Each subplot corresponds to a different test domain: L38 (top-left), L46 (bottom-left), L43 (top-right), and L100 (bottom-right). 
    }
\end{subfigure}
\caption{
Comparison of accuracy between MDG and PMDG settings under equal training data conditions. 
The x-axis shows the total number of training samples, while the y-axis shows the accuracy. 
Blue and orange lines represent MDG and PMDG settings, respectively. 
% Each subplot corresponds to a different test domain: LabelMe (top-left), VOC2007 (bottom-left), SUN09 (top-right), and Caltech101 (bottom-right).
}
\label{fig:same_num}
\Description{
Comparison of accuracy between MDG and PMDG settings under equal training data conditions. 
The x-axis shows the total number of training samples, while the y-axis shows the accuracy. 
Blue and orange lines represent MDG and PMDG settings, respectively. 
}
\end{figure*}

We conducted a comparison between PMDG and conventional MDG approaches under controlled data conditions to assess whether artificially generated pseudo-domains can serve as effective substitutes for naturally occurring domain variations. 
To ensure a fair comparison, we set the total number of training samples equal in both settings. 
Given $n$ samples from a single source domain in the PMDG setting, we constructed an MDG training set by collecting $n/3$ samples from each of three distinct domains, maintaining the same total size $n$. 
These source domain samples were transformed using our Org+IM+IM pseudo-domain generation method in the PMDG setting, while the MDG setting used the original samples directly. 
Both approaches used SD as the base training algorithm for domain generalization.
The experimental results, shown in \Cref{fig:same_num}, revealed several key findings. 
First, the performance of MDG consistently improves with increasing training data size across all datasets, demonstrating a proportional relationship between data size and generalization capability. 
In contrast, PMDG's performance do not show such consistent scaling with data size, which can be attributed to the variations in the source domains used for training.
The effectiveness of PMDG shows a significant dependence on the choice of source domain, with performance varying considerably across different source domain selections. 
Notably, in certain source domain configurations, PMDG achieves superior performance compared to MDG, suggesting that, under optimal conditions, artificially generated domains can be more effective than naturally occurring domain variations.
These findings have important implications for domain generalization research. 
The variable performance of PMDG across different source domains, coupled with its potential to outperform MDG in certain configurations, suggests that the effectiveness of synthetic domain generation is highly dependent on the characteristics of the source domain. 
This indicates that the careful selection of source domains is crucial for the successful implementation of PMDG approaches.
The fact that PMDG can achieve superior performance with specific source domains is particularly promising, as it suggests that when properly configured, artificial domain generation could provide a more effective alternative to traditional multi-domain data collection.
Furthermore, these results challenge the conventional assumption that more diverse training data leads to better generalization. 
While this holds true for MDG approaches, the non-linear relationship between data size and performance in PMDG settings indicates that the quality and characteristics of the source domain may be more important than the quantity of training data.
This insight could lead to more efficient domain generalization strategies that focus on selecting optimal source domains rather than simply collecting more diverse training data.
These observations point to several promising directions for future research, including the development of methods to identify optimal source domains for PMDG and techniques to enhance the effectiveness of pseudo-domain generation across a broader range of source domains. 
Such advances could further establish PMDG as a practical and efficient alternative to traditional MDG approaches, particularly in scenarios where collecting diverse multi-domain data is challenging or resource-intensive.

\subsection{Evaluation on ImageNet}
In our previous experiments, we established the effectiveness of combining MDG algorithms with pseudo-domains.
We then investigated whether this insight could benefit existing SDG research on large-scale datasets. 
Specifically, we evaluated our approach on ImageNet, a standard benchmark in SDG research that offers various distribution shifts through its variants~\citep{imagenet,in-a,in-c,in-c-bar,in-r,in-sketch,in-v2,sty-in}. 
We utilized SD as the MDG algorithm with pseudo-domains Org+IM+IM, comparing it against the baseline SDG method IPMix. 
The selection of IPMix as our baseline is particularly relevant as it represents a state-of-the-art SDG approach.
The results, presented in \Cref{tab:in}, demonstrate the broad applicability of our findings. 
Our method achieves better average accuracy across out-of-distribution variants than the IPMix baseline, showing improvements in generalization performance. 
These results offer several important implications for the SDG research community. 
First, they validate that insights from MDG can indeed enhance existing SDG methods, even on large-scale datasets. 
Second, the performance improvements across multiple distribution shifts suggest that our approach can effectively capture robust features. 
Finally, these findings open new possibilities for advancing SDG research by incorporating established MDG techniques, potentially bridging the gap between these two traditionally separate research directions.

\begin{table*}[tb!]
\centering
\caption{
Accuracy comparison of IPMix and our method on ImageNet (IN) and its variants. 
The highest accuracy for each dataset is shown in \textbf{bold}. 
``Average'' represents the mean accuracy across all test datasets, and ``OOD Average'' denotes the mean accuracy on out-of-distribution test datasets.
}
\label{tab:in}
\begin{tabular}{l|cccccccccc}
\toprule
   Method   & IN  & IN-C           & IN-R           & IN-A          & IN-$\overline{\mathrm{C}}$     & IN-Sketch      & IN-V2          & Stylied-IN     & Average        & OOD Average    \\ \midrule
IPMix & \textbf{77.71} & 51.01          & 42.67          & 4.27          & \textbf{52.11} & 31.07          & 65.61          & 11.54          & 42.00          & 36.90          \\
\rowcolor{lightgray} Ours  & 77.31          & \textbf{51.18} & \textbf{43.78} & \textbf{6.13} & 49.71          & \textbf{31.24} & \textbf{65.91} & \textbf{12.44} & \textbf{42.21} & \textbf{37.20} \\ \bottomrule
\end{tabular}
\end{table*}

\section{Discussion}
Based on our experimental results, we present several key insights that have important implications for future research in domain generalization. 
Through our findings, we aim to bridge the gap between SDG and MDG research.

\textbf{Incorporating MDG Advances into SDG Research.}
Our experiments revealed a positive correlation between algorithm performance in MDG and PMDG settings. 
Furthermore, we found that the combination of SD algorithm with pseudo-domains Org+IM+IM achieved the highest performance, surpassing the SDG baseline. 
These results demonstrate that leveraging established MDG algorithms through our PMDG framework can enhance SDG performance. 
The successful application of MDG algorithms in SDG settings suggests that the traditional separation between these fields may have unnecessarily limited SDG research progress. 
Our PMDG framework provides a practical testbed for MDG algorithms, effectively bridging the gap between advances in MDG and their practical application in SDG scenarios.

\textbf{Reconsidering the Role of SDG Research.}
Our experimental results revealed two important findings. First, pseudo-domains can sometimes outperform actual multi-domains when sufficient training data is available, and MDG algorithms work effectively as learning algorithms in the SDG setting.
These findings suggest that the quality of pseudo-domain generation may have a greater impact on generalization performance than the development of new learning algorithms, as existing MDG algorithms already provide strong learning capabilities. 
Furthermore, our comparison between MDG and PMDG under equal data conditions revealed that the performance gap between them diminishes as training data increases, with PMDG even showing superiority in some cases. 
This observation challenges the conventional assumption that real multi-domain data is always preferable and suggests that with sufficient data, well-designed pseudo-domain generation might be more effective than collecting actual multi-domain datasets.
Consequently, we argue that future SDG research should prioritize the development of better pseudo-domain generation techniques rather than creating new learning algorithms in isolation from MDG advances. 
This shift in focus could potentially lead to more substantial improvements in generalization performance.

\textbf{Future Directions.}
A key direction for future research is the theoretical analysis of when and why pseudo-domains can substitute for actual domains.
This analysis could provide insights into the fundamental principles of domain generalization and guide the development of more effective pseudo-domain generation techniques.
The success of our PMDG framework demonstrates that the artificial boundary between SDG and MDG research has been limiting progress in both fields. 
By providing a bridge between these traditionally separate areas, our work suggests that future advances in domain generalization may come from their synergistic combination: utilizing sophisticated MDG algorithms while focusing SDG research efforts on improving pseudo-domain generation techniques.

\section{Limitation}
A key limitation of our current PMDG framework is its underlying assumption that all transformed data represents distributions distinct from the source data distribution. 
This limitation becomes particularly problematic when dealing with varying transformation intensities, as weakly transformed data may remain substantially similar to the source distribution. 
Consequently, treating such transformed data as distinct domains could lead to suboptimal performance by artificially emphasizing minor variations that do not contribute to meaningful domain generalization.
While addressing this limitation would require extensive theoretical analysis and empirical validation, a potential solution would be to quantitatively assess the relationship between source and transformed distributions. 
% This could involve measuring distribution distances or utilizing domain classifiers to determine if a transformation creates a meaningful shift from the source domain.  
Such analysis, although computationally intensive, could provide a more principled foundation for pseudo-domain generation and potentially lead to more effective domain generalization strategies.

\section{Conclusion}
In this work, we proposed the Pseudo Multi-source Domain Generalization (PMDG) framework to bridge the gap between Single-Domain Generalization (SDG) and Multi-Domain Generalization (MDG) research. 
Through extensive experiments, we demonstrated that incorporating MDG algorithms into SDG settings via pseudo-domain generation can enhance generalization performance.
We obtained several important findings from our experiments. 
First, MDG algorithms can be effectively utilized in SDG settings through our PMDG framework, with the combination of ViT backbone, SD algorithm and Org+ST+ED+CG+IM+IM achieving state-of-the-art performance. 
Second, when sufficient training data is available, pseudo-domains can serve as effective substitutes for actual multi-source domains, suggesting that future SDG research should focus on developing better pseudo-domain generation techniques rather than new learning algorithms. 
Third, the effectiveness of pseudo-domain generation techniques can vary with network architectures, as demonstrated by our experiments with ResNet50 and ViT.
Our work also reveals important considerations for domain generalization research. 
The dataset-specific effectiveness of certain pseudo-domain combinations (\eg, style-based transformations for PACS) highlights the importance of evaluating methods across diverse distribution shifts. 
These findings open new directions for future research, particularly in understanding the theoretical relationship between pseudo-domains and actual domains. 
By providing a bridge between SDG and MDG research, our work suggests that future advances in domain generalization may come from their synergistic combination rather than treating them as separate fields.

\clearpage

\bibliographystyle{ACM-Reference-Format}
\bibliography{main}

\clearpage
\appendix
\setcounter{table}{2}
\setcounter{figure}{5}
\setcounter{equation}{3}

In this Supplementary Material, we provide additional analysis for the PMDG. 

% \section{Implementation Details of Pseudo-domain Generation in PseudoDomainBed}
% Implementing various transformation techniques in a unified framework presents challenges due to their different operating levels. 
% To address this, we carefully designed the implementation architecture of PseudoDomainBed to handle different types of transformations consistently.

% Image transformations in our framework can be categorized into two levels based on their processing stage.
% The first category is dataset-level transformations, which operate on raw images before converting them to tensors. 
% Transformations such as RandAugment fall into this category, where augmentations are applied directly to image data.
% The second category is mini-batch-level transformations, which operate on normalized tensors during the training process. 
% For example, MixUp belongs to this category as it combines multiple normalized image tensors.

% To handle these different transformation types uniformly, we implemented a standardized interface for pseudo-domain generation algorithms. 
% Each algorithm is required to implement both dataset-level and mini-batch-level transformation methods, even if only one is actually used. 
% This design choice provides a consistent API for users to employ any transformation technique without considering its implementation level. 
% It also enables flexible integration of new transformation methods by implementing the standard interface. 

\section{Extended Analysis of Pseudo-domain Combinations}
While the main paper focuses on two specific pseudo-domain combinations, here we present a comprehensive evaluation of various combinations to better understand their effectiveness. 
We evaluated these combinations using ERM and SD as the MDG algorithms, as they demonstrated strong performance in our main experiments.
\Cref{tab:combination} presents the results of this extended analysis. 
The experimental results revealed several important insights about pseudo-domain combinations. 
First, not all combinations lead to improved performance, with some degrading accuracy compared to using no pseudo-domains. 
This negative effect is particularly pronounced when incorporating CutMix into the combinations, suggesting that certain transformations may interfere with the model's ability to learn robust features.
Second, we found that applying IPMix multiple times consistently outperforms combinations of different data augmentation techniques. 
This suggests that the quality and consistency of the pseudo-domain generation technique may be more important than the diversity of transformation types.
Third, while style-based transformations alone show moderate effectiveness, their combination with IPMix leads to further improvements in performance. 
This complementary effect indicates that style-based transformations and IPMix capture different aspects of domain variation, making their combination particularly effective for domain generalization.

\begin{table*}[tb!]
\centering
\caption{
Accuracy comparison of different pseudo-domain combinations on four datasets. 
``Avg'' represents the mean accuracy across all datasets. 
In each table, the upper part shows SDG baseline results and the lower part presents PMDG results. 
% \textbf{Bold} and \underline{underlined} numbers denote the first and second highest accuracy, respectively. 
Org denotes the original domain without transformation, IM represents pseudo-domains generated by IPMix, AM by AugMix, MU by Mixup, CM by CutMix, RC by RandConv, ST by StyleTransfer, ED by EdgeDetection, and CG by CartoonGAN.
$^\dagger$ indicates exclusion of domain-specific transformations during training: ST is excluded when testing on Art domain, CG for Cartoon domain, and ED for Sketch domain.
}
\label{tab:combination}
\resizebox{\textwidth}{!}{
\begin{tabular}{llccccc}
\toprule
\textbf{Algorithm} & \textbf{Pseudo-domain}             & \textbf{VLCS}  & \textbf{PACS}  & \textbf{OfficeHome} & \textbf{TerraIncognita} & \textbf{Avg} \\ \midrule
ERM                & --                                 & 60.6 $\pm$ 1.3 & 56.3 $\pm$ 0.5 & 53.4 $\pm$ 0.1      & 32.2 $\pm$ 0.7          & 50.6         \\
Mixup              & --                                 & 64.3 $\pm$ 1.2 & 55.9 $\pm$ 0.8 & 54.5 $\pm$ 0.3      & 32.7 $\pm$ 0.5          & 51.8         \\
SagNet             & --                                 & 62.0 $\pm$ 0.3 & 52.2 $\pm$ 0.4 & 51.5 $\pm$ 0.3      & 32.6 $\pm$ 0.7          & 49.6         \\
RSC                & --                                 & 65.0 $\pm$ 0.9 & 56.1 $\pm$ 1.3 & 52.5 $\pm$ 0.3      & 33.3 $\pm$ 0.2          & 51.7         \\
AugMix             & --                                 & 61.8 $\pm$ 0.5 & 57.8 $\pm$ 0.4 & 54.6 $\pm$ 0.3      & 32.1 $\pm$ 0.3          & 51.6         \\
CutMix             & --                                 & 61.8 $\pm$ 0.3 & 54.7 $\pm$ 1.0 & 53.9 $\pm$ 0.1      & 33.0 $\pm$ 1.1          & 50.8         \\
IPMix              & --                                 & 64.6 $\pm$ 1.0 & 65.9 $\pm$ 0.3 & 55.6 $\pm$ 0.2      & 34.9 $\pm$ 0.7          & 55.2         \\
RandAugment        & --                                 & 58.6 $\pm$ 0.8 & 58.9 $\pm$ 1.0 & 53.9 $\pm$ 0.2      & 33.2 $\pm$ 0.5          & 51.1         \\
RandConv           & --                                 & 62.1 $\pm$ 0.1 & 62.8 $\pm$ 0.7 & 53.2 $\pm$ 0.3      & 34.7 $\pm$ 0.3          & 53.2         \\
TrivialAugment     & --                                 & 61.1 $\pm$ 1.1 & 59.9 $\pm$ 1.5 & 54.2 $\pm$ 0.2      & 36.2 $\pm$ 0.2          & 52.8         \\ 
\midrule
ERM                & Org+AM+MU+CM       & 59.0 $\pm$ 0.6 & 51.6 $\pm$ 0.6 & 51.8 $\pm$ 0.2      & 34.9 $\pm$ 0.2          & 49.4         \\
ERM                & Org+IM+IM               & 64.6 $\pm$ 1.5 & 63.4 $\pm$ 1.3 & 55.1 $\pm$ 0.4      & 36.6 $\pm$ 0.9          & 54.9         \\
ERM                & Org+IM+RC            & 65.0 $\pm$ 0.5 & 61.3 $\pm$ 0.6 & 55.9 $\pm$ 0.3      & 33.8 $\pm$ 0.9          & 54.0         \\
ERM & Org+ST+ED+CT$^\dagger$ & 60.2 $\pm$ 0.7            & 59.7                      & 55.3                      & 28.2 $\pm$ 0.7            & 50.9                     \\
% ERM & Org+ST+ED+CT+IM        & 60.1 $\pm$ 0.8            & 73.2 $\pm$ 0.4            & 55.5 $\pm$ 0.2            & 31.5 $\pm$ 0.9            & 55.1                      \\
% ERM & Org+ST+ED+CT+IM+RC     & 59.3 $\pm$ 0.3            & 73.2 $\pm$ 0.3            & 55.4 $\pm$ 0.2            & 30.6 $\pm$ 0.4            & 54.6                      \\
% ERM & Org+ST+ED+CT+RC        & 61.6 $\pm$ 0.4            & 72.8 $\pm$ 0.2            & 56.2 $\pm$ 0.1            & 27.9 $\pm$ 0.6            & 54.6                      \\
ERM                & Org+ST+ED+CG+IM+IM$^\dagger$     & 64.9 $\pm$ 1.1 & 69.9 $\pm$ 0.5 & 55.4 $\pm$ 0.1 & 31.1 $\pm$ 0.6  & 55.3\\
SD                 & AM+IM                       & 63.8 $\pm$ 0.6 & 63.5 $\pm$ 1.5 & 55.5 $\pm$ 0.3      & 37.5 $\pm$ 0.6          & 55.1         \\
SD                 & MU+IM                        & 62.2 $\pm$ 0.6 & 54.1 $\pm$ 1.0 & 54.7 $\pm$ 0.2      & 36.8 $\pm$ 0.7          & 51.9         \\
SD                 & Org+AM+IM              & 65.4 $\pm$ 1.2 & 62.1 $\pm$ 0.5 & 56.3 $\pm$ 0.4      & 37.0 $\pm$ 0.2          & 55.2         \\
SD                 & Org+AM+MU              & 62.0 $\pm$ 1.0 & 57.1 $\pm$ 0.9 & 54.3 $\pm$ 0.3      & 37.3 $\pm$ 1.1          & 52.7         \\
SD                 & Org+AM+MU+CM       & 60.9 $\pm$ 0.9 & 54.5 $\pm$ 0.7 & 52.4 $\pm$ 0.4      & 35.4 $\pm$ 0.5          & 50.8         \\
SD                 & Org+AM+MU+CM+IM & 62.0 $\pm$ 0.3 & 55.3 $\pm$ 0.4 & 53.8 $\pm$ 0.9      & 34.7 $\pm$ 1.1          & 51.5         \\
SD                 & Org+MU+IM               & 65.1 $\pm$ 1.1 & 58.9 $\pm$ 0.6 & 55.0 $\pm$ 0.2      & 38.7 $\pm$ 0.9          & 54.4         \\
SD                 & Org+IM+RC            & 63.7 $\pm$ 0.7 & 63.1 $\pm$ 0.7 & 56.6 $\pm$ 0.2      & 36.8 $\pm$ 0.5          & 55.1         \\
SD                 & Org+IM                     & 62.8 $\pm$ 1.3 & 62.3 $\pm$ 0.5 & 56.3 $\pm$ 0.1      & 37.2 $\pm$ 1.3          & 54.7         \\
SD                 & Org+IM+IM               & 65.6 $\pm$ 1.2 & 64.1 $\pm$ 1.0 & 56.7 $\pm$ 0.2      & 37.1 $\pm$ 0.7          & 55.9         \\
SD                 & Org+IM+IM+IM         & 66.0 $\pm$ 1.1 & 61.0 $\pm$ 0.6 & 56.8 $\pm$ 0.1      & 37.4 $\pm$ 0.1          & 55.3         \\
SD & Org+ST+ED+CT$^\dagger$  & 61.3 $\pm$ 1.0            & 60.5                       & 56.1                     & 30.0 $\pm$ 0.4            & 52.0                      \\
% SD & Org+ST+ED+CT+IM         & 60.7 $\pm$ 0.8            & 73.3 $\pm$ 0.1            & 56.7 $\pm$ 0.2            & 32.8 $\pm$ 0.6            & 55.9                      \\
% SD & Org+ST+ED+CT+IM+RC      & 59.8 $\pm$ 1.2            & 74.3 $\pm$ 0.0            & 56.6 $\pm$ 0.2            & 31.8 $\pm$ 0.6            & 55.6                      \\
% SD & Org+ST+ED+CT+RC         & 60.9 $\pm$ 0.4            & 73.5 $\pm$ 0.3            & 56.6 $\pm$ 0.2            & 31.3 $\pm$ 0.5            & 55.6                      \\
SD                 & Org+ST+ED+CG+IM+IM$^\dagger$    & 61.4 $\pm$ 0.3 & 69.7 $\pm$ 0.4         & 55.6 $\pm$ 0.2          & 30.5 $\pm$ 0.9 & 54.3 \\
\bottomrule
\end{tabular}
}
\end{table*}

\section{Additional Visualization Results}
We present more visualization results in \Cref{fig:image_transformations}, including a comprehensive comparison of various data augmentation and transformation techniques. 
The figure shows the visual effects of different approaches: No Data Augmentation as a baseline, Default Data Augmentation as a standard practice on DomainBed, and ten different transformation techniques including AugMix, CartoonGAN, CutMix, Edge Detection, IPMix, MixUp, RandAugment, RandConv, Style Transfer, and TrivialAugment. 
These visualizations provide insights into how each technique modifies the input images, offering a clearer understanding of their distinctive characteristics and potential impact on model training.

\begin{figure*}[tb!]
\centering
\begin{subfigure}[tb!]{0.45\textwidth}
    \includegraphics[width=\textwidth]{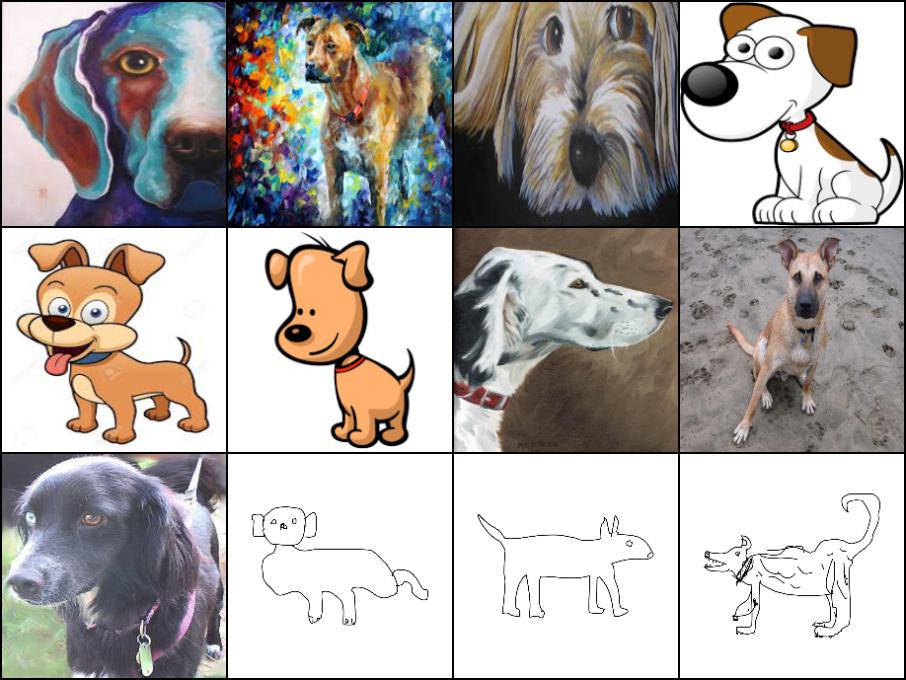}
    \caption{No Data Augmentation}
\end{subfigure}
\begin{subfigure}[tb!]{0.45\textwidth}
    \includegraphics[width=\textwidth]{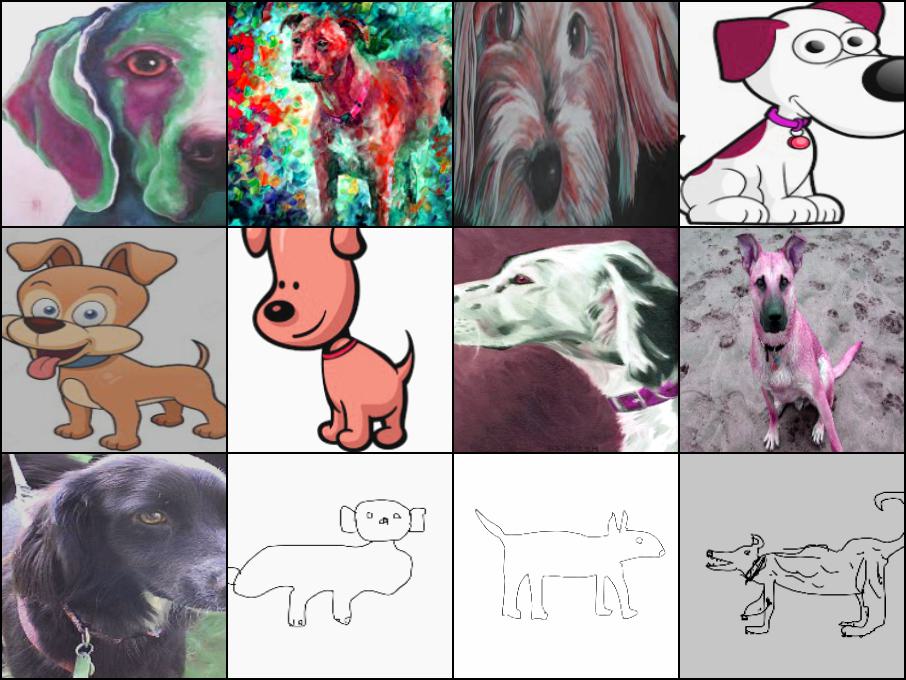}
    \caption{Default Data Augmentation}
\end{subfigure}

\begin{subfigure}[tb!]{0.45\textwidth}
    \includegraphics[width=\textwidth]{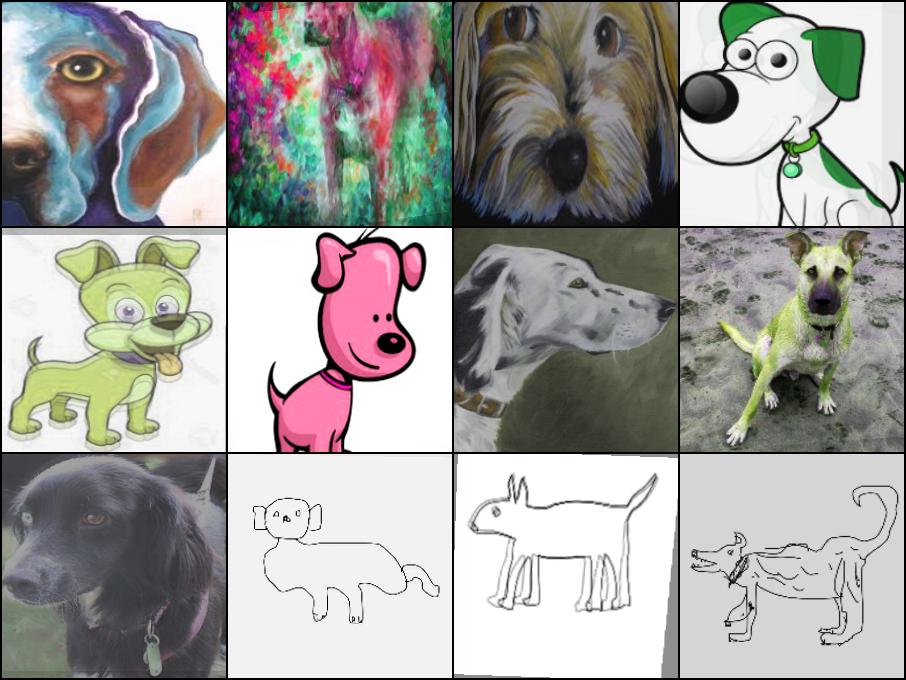}
    \caption{AugMix}
\end{subfigure}
\begin{subfigure}[tb!]{0.45\textwidth}
    \includegraphics[width=\textwidth]{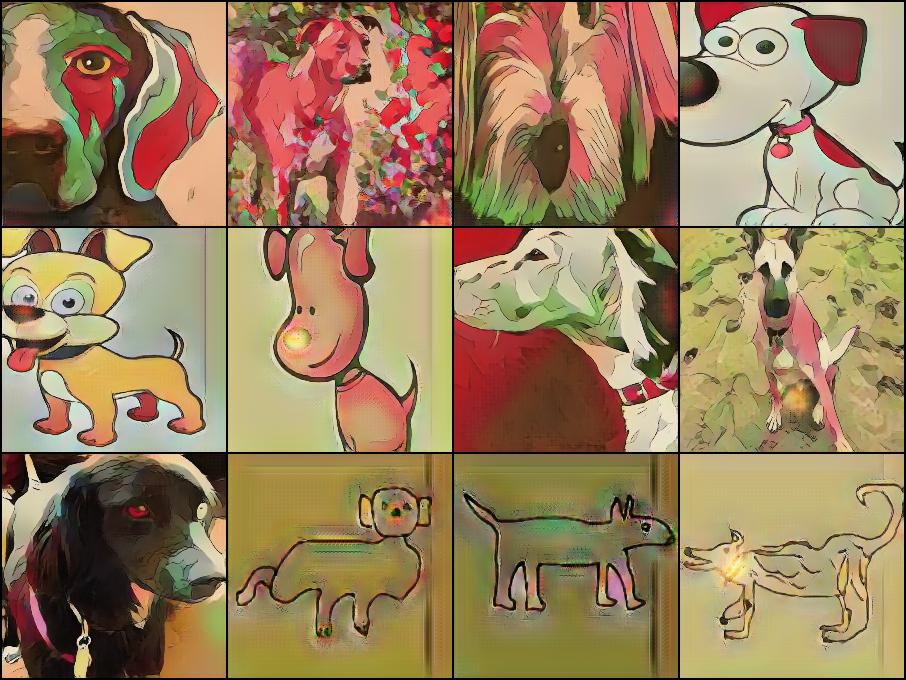}
    \caption{CartoonGAN}
\end{subfigure}

\begin{subfigure}[tb!]{0.45\textwidth}
    \includegraphics[width=\textwidth]{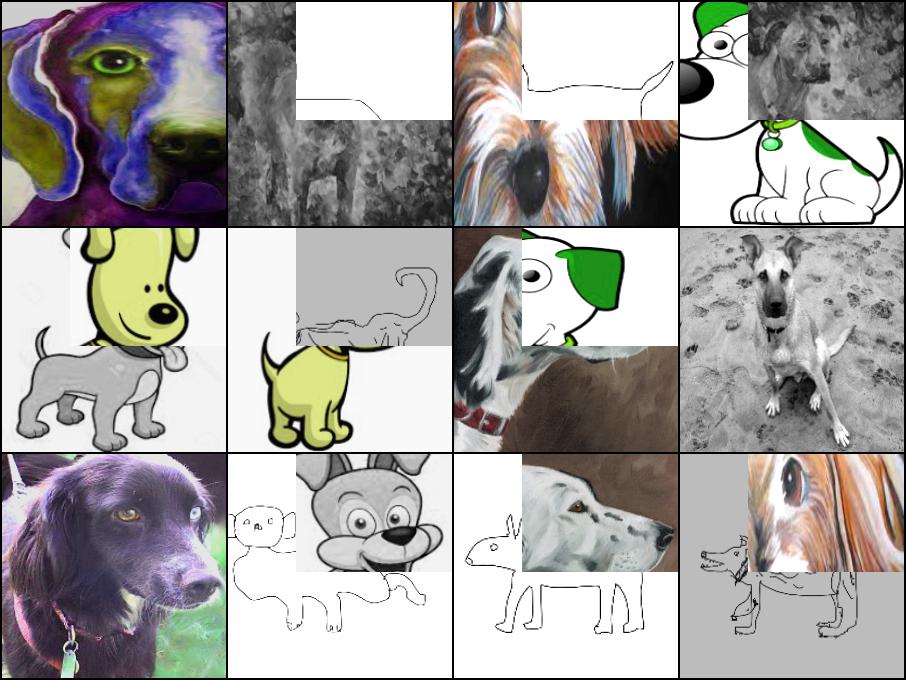}
    \caption{CutMix}
\end{subfigure}
\begin{subfigure}[tb!]{0.45\textwidth}
    \includegraphics[width=\textwidth]{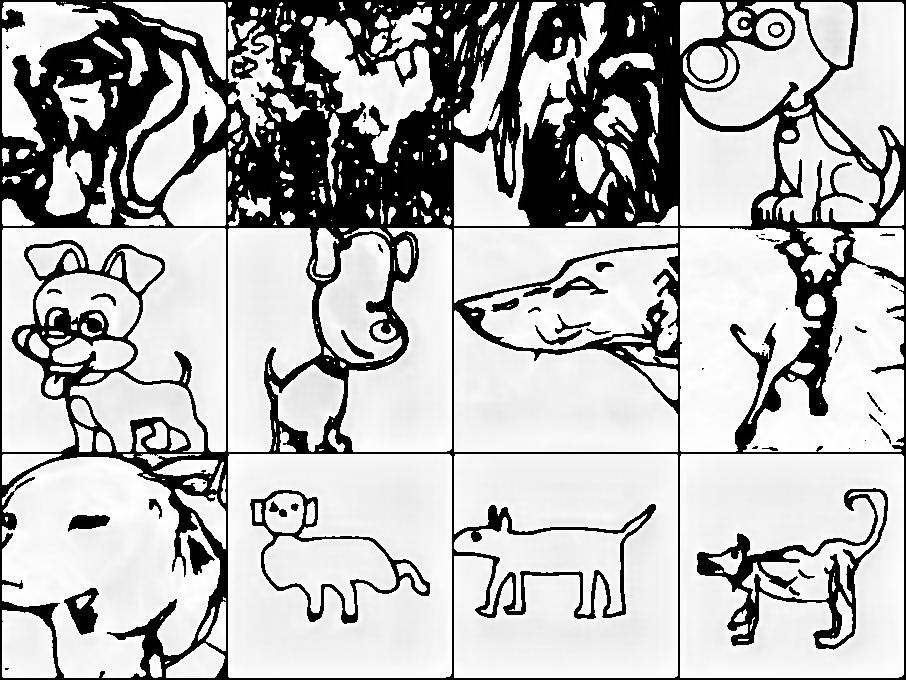}
    \caption{Edge Detection}
\end{subfigure}
\end{figure*}

\begin{figure*}[p]
\ContinuedFloat
\centering
\begin{subfigure}[tb!]{0.45\textwidth}
    \includegraphics[width=\textwidth]{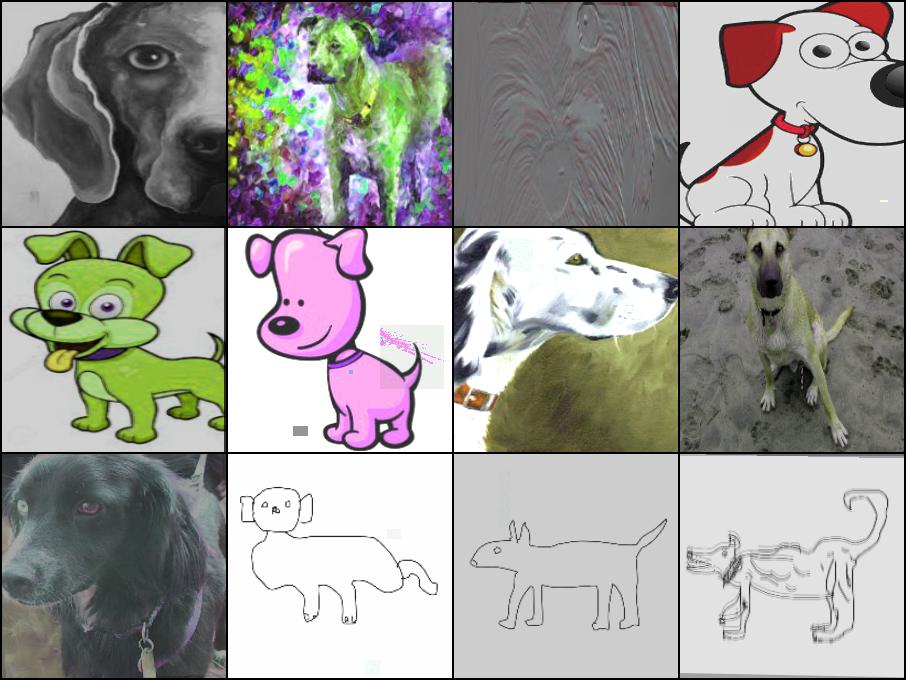}
    \caption{IPMix}
\end{subfigure}
\begin{subfigure}[tb!]{0.45\textwidth}
    \includegraphics[width=\textwidth]{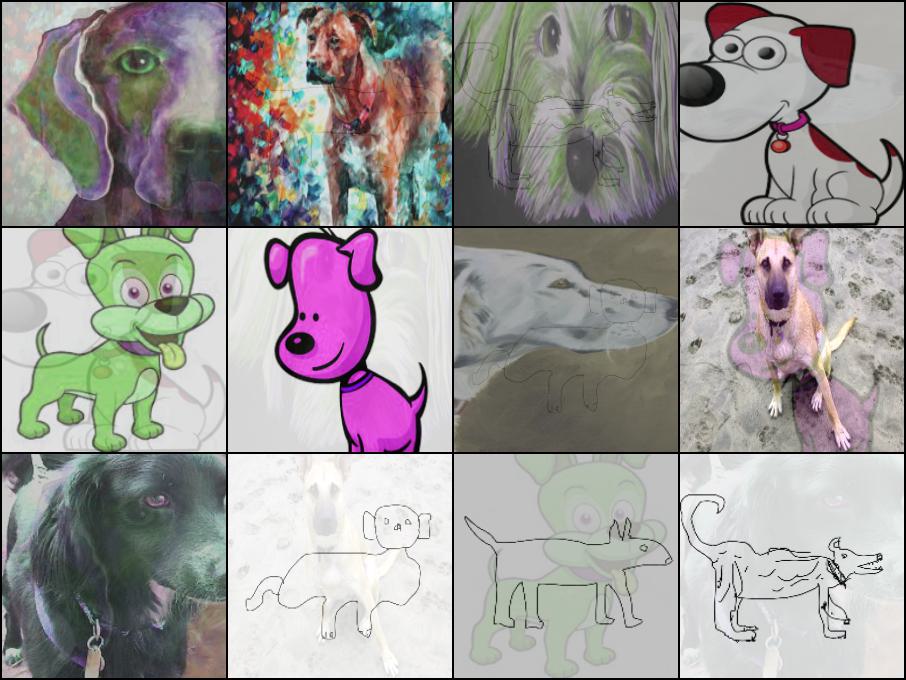}
    \caption{MixUp}
\end{subfigure}

\begin{subfigure}[tb!]{0.45\textwidth}
    \includegraphics[width=\textwidth]{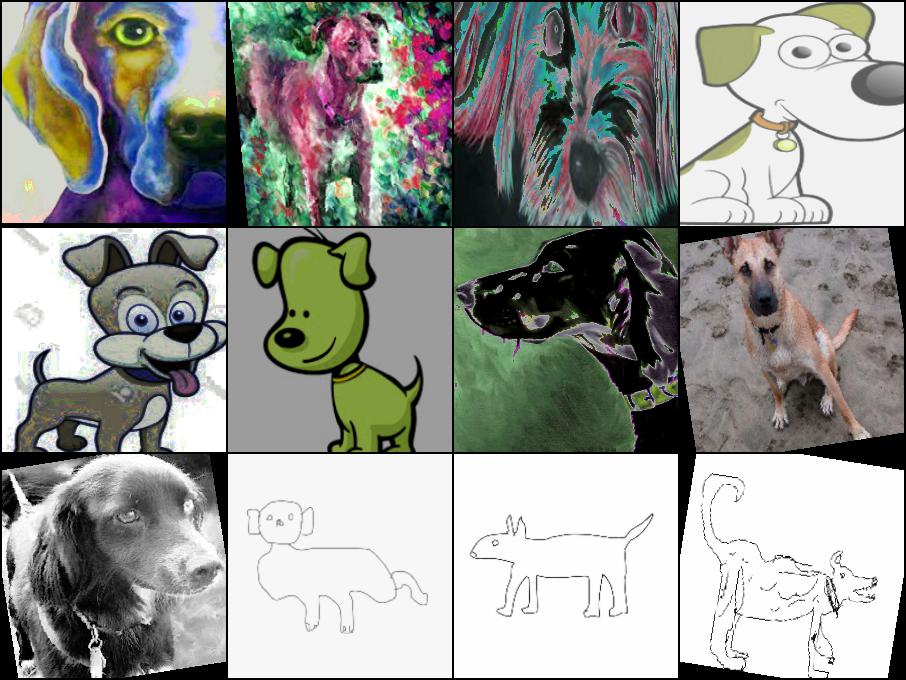}
    \caption{RandAugment}
\end{subfigure}
\begin{subfigure}[tb!]{0.45\textwidth}
    \includegraphics[width=\textwidth]{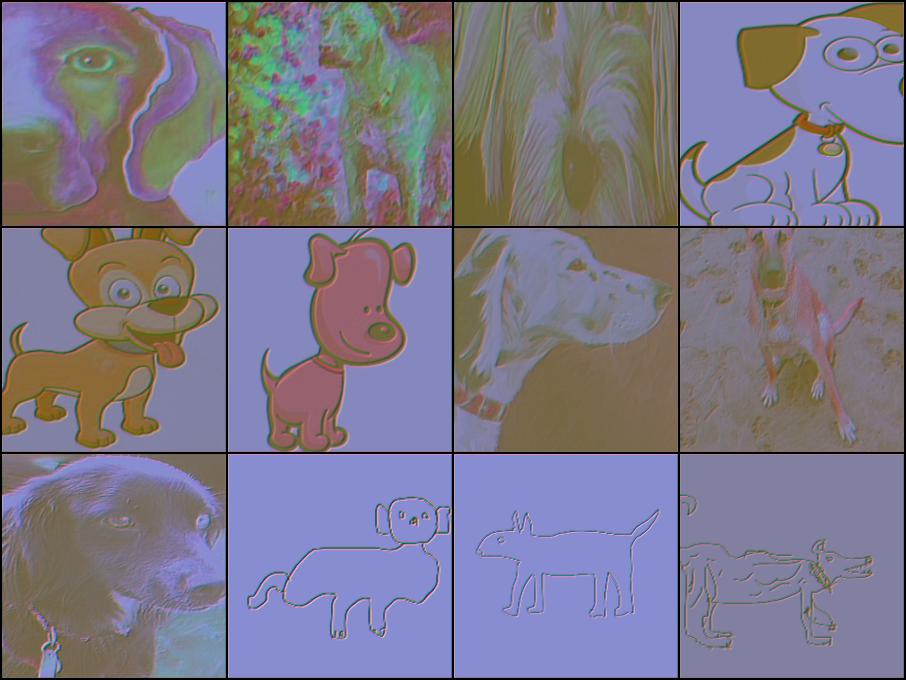}
    \caption{RandConv}
\end{subfigure}

\begin{subfigure}[tb!]{0.45\textwidth}
    \includegraphics[width=\textwidth]{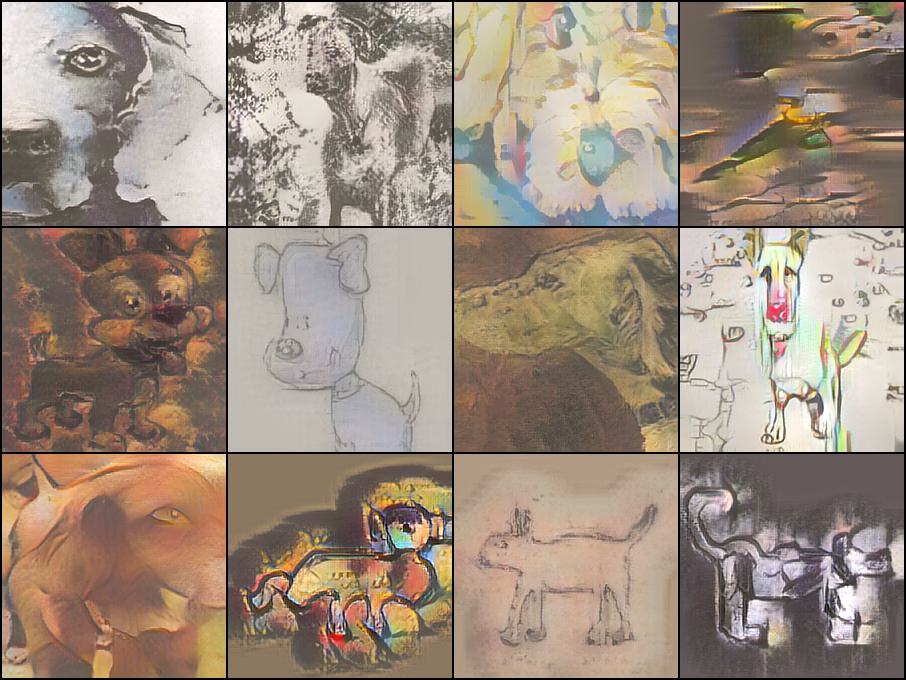}
    \caption{Style Transfer}
\end{subfigure}
\begin{subfigure}[tb!]{0.45\textwidth}
    \includegraphics[width=\textwidth]{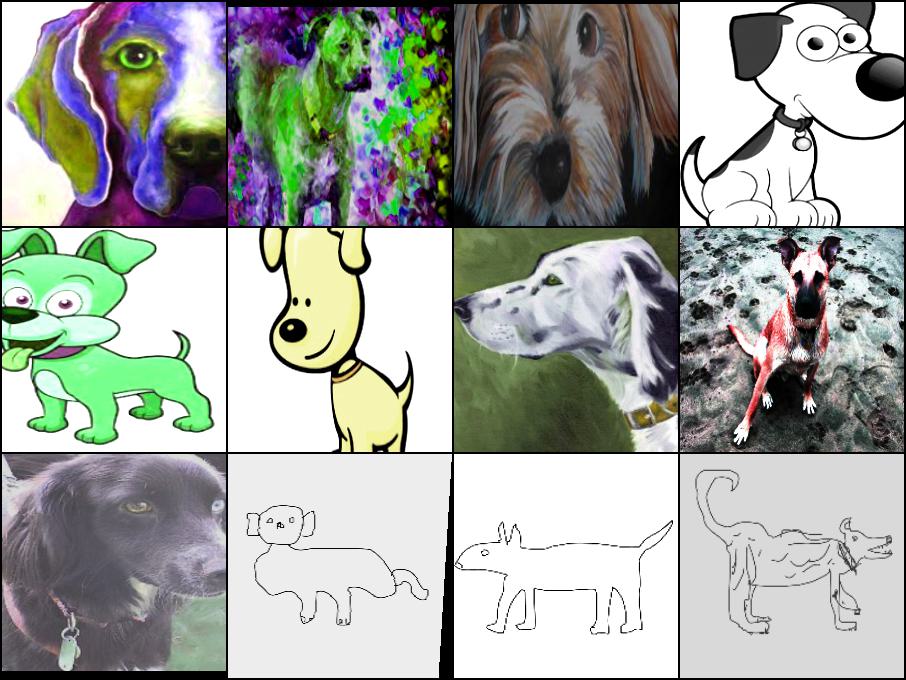}
    \caption{TrivialAugment}
\end{subfigure}

\caption{
Comparison of Various Image Transformation Techniques. 
Default Data Augmentation refers to the standard data augmentation settings, while No Data Augmentation indicates the case where no data augmentation was applied.
}
\label{fig:image_transformations}
\end{figure*}

\end{document}